\documentclass[journal]{IEEEtran}

\ifCLASSINFOpdf
\else
   \usepackage[dvips]{graphicx}
\fi
\usepackage{url}

\usepackage{graphicx}
\usepackage{amssymb} 
\usepackage{booktabs} 
\usepackage{graphicx} 
\usepackage{array} 
\usepackage{amsmath} 
\usepackage{hyperref}
\usepackage{caption}
\usepackage{xcolor}

\begin{document}

\title{CLIP-TNseg: A Multi-Modal Hybrid Framework for Thyroid Nodule Segmentation in Ultrasound Images}

\author{Xinjie Sun, Boxiong Wei, Yalong Jiang, \IEEEmembership{Member, IEEE}, Liquan Mao, and Qi Zhao, \IEEEmembership{Member, IEEE} 
\thanks{Manuscript received 2 December 2024. This work was supported by the National Natural Science Foundation of China under Grant 62301020 and by Beijing Natural Science Foundation under Grant 4234085.}
\thanks{Xinjie Sun, Liquan Mao, Qi Zhao are with the School of Electronic and Information, Beihang University, Beijing 100191, China  (e-mail: jayxjsun@gmail.com; e-mail: findletliquanmao@outlook.com; e-mail: zhaoqibuaa@gmail.com).}
\thanks{Boxiong Wei is with the department of Ultrasound, Peking University First Hospital, Beijing, China (e-mail: weiboxiong123@126.com).}

\thanks{Yalong Jiang (corresponding author) is with Unmanned
System Research Institute, Beihang University, Beijing 100191, China  (e-mail: AllenYLJiang@outlook.com).}

}

\markboth{IEEE SIGNAL PROCESSING LETTERS, Vol. 14, No. 8, August 2015}
{Shell \MakeLowercase{\textit{et al.}}: Bare Demo of IEEEtran.cls for IEEE Journals}
\maketitle

\begin{abstract}
Thyroid nodule segmentation in ultrasound images is crucial for accurate diagnosis and treatment planning. However, existing methods face challenges in segmentation accuracy, interpretability, and generalization, which hinder their performance. This letter proposes a novel framework, CLIP-TNseg, to address these issues by integrating a multimodal large model with a neural network architecture. CLIP-TNseg consists of two main branches: the Coarse-grained Branch, which extracts high-level semantic features from a frozen CLIP model, and the Fine-grained Branch, which captures fine-grained features using U-Net style residual blocks. These features are fused and processed by the prediction head to generate precise segmentation maps. CLIP-TNseg leverages the Coarse-grained Branch to enhance semantic understanding through textual and high-level visual features, while the Fine-grained Branch refines spatial details, enabling precise and robust segmentation. Extensive experiments on public and our newly collected datasets demonstrate its competitive performance. Our code and the original dataset are available at https://github.com/jayxjsun/CLIP-TNseg.
\end{abstract}


\IEEEpeerreviewmaketitle

\begin{figure*}[!ht]
	\centering
	\begin{center}
		\includegraphics[width=1.0 \textwidth]{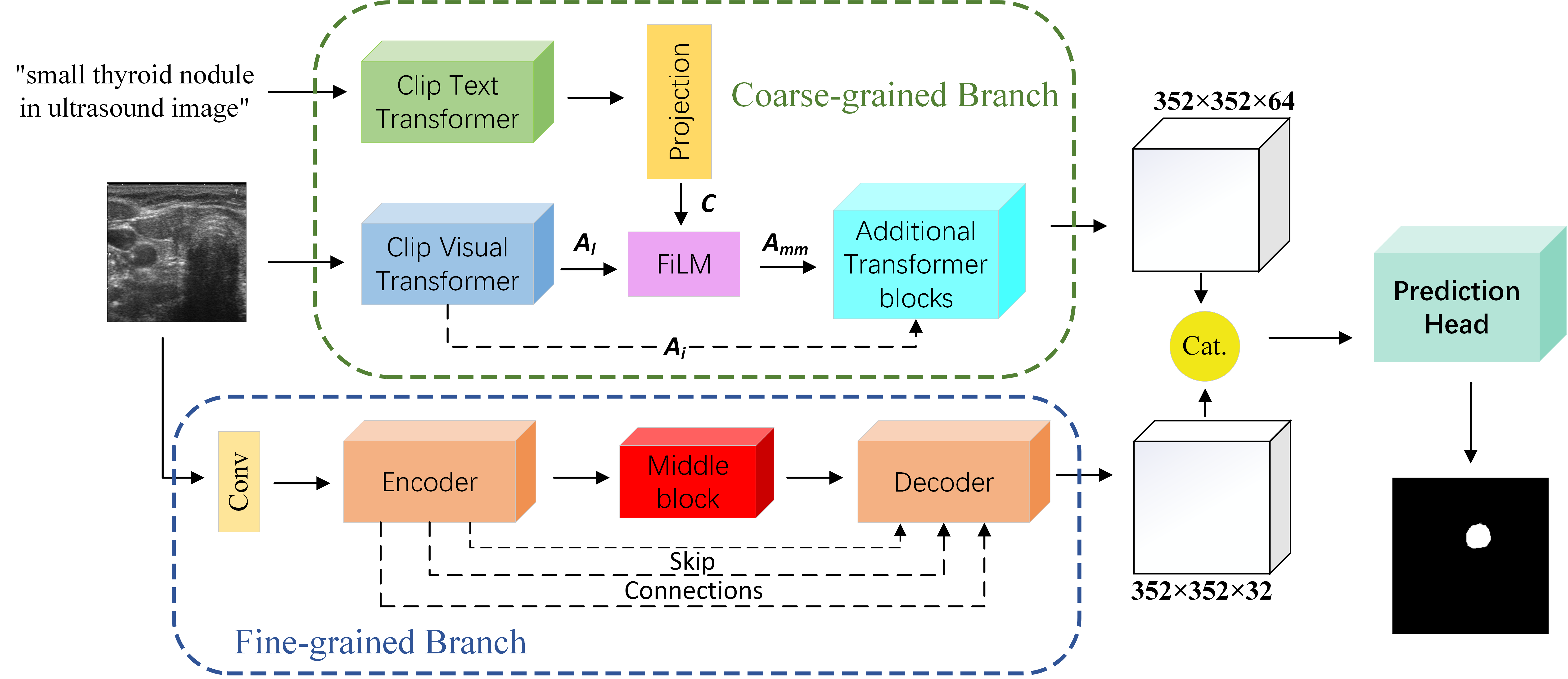}
	\end{center}
    
	\caption{Overall architecture of our CLIP-TNseg, with three main components: CGB for extracting high-level semantic features from pre-trained CLIP models, FGB for capturing fine-grained features with residual learning, and PH for generating final segmentation maps. } 
	\label{fig:fig1}
	\vspace{-8pt}
\end{figure*}

\section{Introduction}

\IEEEPARstart{T}{hyroid} nodules, detectable in up to 68\% of the adult population with high-resolution ultrasound\cite{guth2009very}, are mostly benign, but approximately 5-15\% have the potential to be malignant, highlighting the importance of early detection and precise evaluation\cite{cooperds2006management}. Ultrasound is the preferred imaging method for assessing thyroid nodules due to its non-invasive nature, real-time imaging capabilities, and affordability\cite{tessler2017acr}. Accurate segmentation of thyroid nodules in ultrasound images is essential for proper diagnosis, monitoring, and treatment, as it allows clinicians to measure the size, shape, and structure of the nodules\cite{wang2020automatic}. However, ultrasound image quality is often compromised by speckle noise, low contrast, and artifacts like shadowing, making it challenging to clearly delineate the boundaries of the nodules\cite{noble2006ultrasound}. This highlights the need for more advanced techniques to improve segmentation performance in clinical settings\cite{wang2021automatic}.

Deep learning has enabled the success of segmentation methods like FCN\cite{long2015fully} and U-Net\cite{ronneberger2015u} in medical image analysis, with variants like Attention U-Net\cite{oktay2018attention} and DMU-Net\cite{yang2022dmu} enhancing feature extraction and precision. However, ultrasound images, with their noisy backgrounds and unclear target structures, pose challenges for these methods. Transformers\cite{vaswani2017attention}, such as ViT\cite{dosovitskiy2020image}, TransUNet\cite{chen2021transunet}, and Swin-UNet\cite{cao2022swin}, attempt to address this by capturing long-range dependencies, but they still struggle with accuracy and interpretability in medical applications\cite{shamshad2023transformers}.

The emergence of multimodal large models offers a promising solution to address the limitations of traditional deep learning methods\cite{radford2021learning}. Unlike conventional approaches, multimodal models integrate diverse sources of information, such as textual descriptions and visual inputs, to enhance segmentation performance. Specifically, the textual modality serves as semantic guidance, helping the model better understand and identify target features\cite{luddecke2022image}. Richer textual information improves segmentation accuracy and enhances model interpretability. Additionally, large models trained on vast amounts of data are capable of learning abstract, high-level features, which can be effectively leveraged with minimal fine-tuning on specific datasets.

To address the challenge of thyroid nodule segmentation in medical ultrasound images, we propose CLIP-TNseg, a novel architecture combining the strengths of a multimodal large model and neural networks. CLIP-TNseg consists of two branches: the Coarse-grained Branch(CGB), which extracts high-level semantic features from the frozen CLIP model, and the Fine-grained Branch(FGB), which captures fine-grained features using residual blocks and convolution operations. These features are concatenated and fed into the Prediction Head to produce the final segmentation map. Finally, we conduct experiments on a large, comprehensive dataset (comprising public datasets and a dataset collected by our team) to evaluate the proposed method.

The contributions of this work are as follows:
\begin{itemize}
\item[$\bullet$] We innovatively propose CLIP-TNseg, which incorporates textual information to assist in constructing robust representations, marking a novel application of the multimodal large model CLIP to thyroid nodule segmentation in ultrasound images.

\item[$\bullet$]We design a novel two-branch architecture for thyroid nodule segmentation, where CGB captures high-level semantic features for the target's primary structure, while FGB focuses on fine-grained features for precise segmentation, ensuring both robust feature extraction and accurate boundary delineation.

\item[$\bullet$] We collect and construct a thyroid nodule segmentation dataset (PKTN) as part of this study, and we validate the effectiveness of our method through comparisons and ablation studies on this dataset, alongside other public datasets.
\end{itemize}

\section{PROPOSED METHOD}

\subsection{Overview}

Fig. \ref{fig:fig1} illustrates the overall architecture of the proposed method, which consists of three main components: the Coarse-grained Branch(CGB), the Fine-grained Branch(FGB), and the prediction head(PH). The proposed framework processes an ultrasound image and a textual description as input, aiming to output a pixel-wise segmentation map. Inspired by \cite{luddecke2022image}, \cite{sanderson2022fcn}, CGB and FGB are responsible for feature extraction from the input samples. In CGB, we integrate textual and visual features using FiLM (Feature-wise Linear Modulation)\cite{dumoulin2018feature}, enabling effective cross-modal feature fusion. Meanwhile, FGB employs U-Net style skip connections to achieve multi-level feature fusion across different layers. After extracting features from both branches, these are concatenated and fed into the prediction head to generate the final segmentation map.

\subsection{Coarse-grained Branch (CGB)}

The Coarse-grained Branch is designed to capture high-level semantic information for understanding the overall structure of segmentation targets. It includes a frozen CLIP visual encoder\cite{radford2021learning} and additional transformer blocks\cite{luddecke2022image}. The frozen encoder (ViT-B/16), pre-trained on large-scale image-text datasets, efficiently extracts semantic features, while the transformer blocks refine these visual features and integrate textual inputs. By leveraging CLIP’s text transformer, this branch enables the model to condition segmentation on textual queries, providing precise and flexible control over the segmentation targets.

Specifically, we extract activations from the CLIP visual encoder to provide multi-scale features for segmentation. While the input image \( \mathbf{X} \in \mathbb{R}^{W \times H \times 3} \) is passed through the CLIP Visual Transformer, activations are extracted from multiple layers $S$. Activations are projected to a fixed embedding size $D$, producing features \( \mathbf{A}_i  \) for each layer \( i \in S \).

To incorporate textual supervision, the CLIP Text Transformer processes a text query to generate a conditional vector \( \mathbf{C} \), which facilitates cross-modal feature fusion through FiLM. The vector \( \mathbf{C} \) influences the scaling and shifting parameters \( \gamma \) and \( \beta \), which are then applied to the feature \( \mathbf{A}_l \), extracted from the last layer in the extraction process. The fusion process is defined as:

\begin{equation}
	\mathbf{A}_{\text{mm}} = \gamma(\mathbf{C}) \odot \mathbf{A}_l + \beta(\mathbf{C}),
\end{equation}
where \( \odot \) denotes element-wise multiplication.

The fused multi-modal features \( \mathbf{A}_{\text{mm}} \) are passed into the first additional transformer block, which processes the feature as:

\begin{equation}
	\mathbf{F}_1 = \text{TransformerBlock}_1(\mathbf{A}_{\text{mm}}).
\end{equation}

Subsequent transformer blocks refine the features iteratively. Specifically, each transformer block \( j \) processes the combination of the previously refined activations \( \mathbf{F}_{j-1} \) and the corresponding extracted feature \( \mathbf{A}_i \) as follows:

\begin{equation}
	\mathbf{F}_j = \text{TransformerBlock}_j(\mathbf{F}_{j-1} + \mathbf{A}_i),
\end{equation}
where \( \mathbf{A}_i \) refers to the extracted activation from the corresponding layer in the CLIP encoder, and the activations are processed in reverse order of extraction. The output tokens from the final transformer block are reshaped via transposed convolution to \( \mathbb{R}^{W \times H \times D} \), preparing them for feature fusion in later stages.

This combination of the frozen CLIP backbone and transformer blocks enables the segmentation model to learn detailed patterns for accurate results. By employing FiLM for cross-modal feature fusion based on textual queries, the model leverages a shared vision-text embedding space, enhancing its adaptability and interpretability. Following CLIPSeg\cite{luddecke2022image} guidelines, we use a patch size of $P=16$ and a projection dimension of $D=64$, extracting activations from layers $S=[3,7,9]$ at an image resolution of 352 pixels.

\subsection{Fine-grained Branch (FGB)}
The Fine-grained Branch (FGB) is designed to extract fine-grained features while maintaining multi-scale feature fusion, inspired by the U-Net architecture \cite{ronneberger2015u}. It takes the input image and leverages residual blocks \cite{he2016deep} along with downsampling and upsampling operations to effectively extract and reconstruct features.

In the encoder, features are progressively extracted through a series of residual blocks, capturing rich representations at multiple scales. The decoder employs upsampling and skip connections to fuse high-resolution encoder features with decoder features, recovering spatial details for accurate segmentation.

By combining residual learning and U-Net-style skip connections, FGB complements high-level features from CGB with fine-grained feature map outputs, improving segmentation accuracy by refining spatial details.

\subsection{Prediction Head (PH)}
The Prediction Head (PH) takes the full-size tensor, formed by concatenating the feature maps from both CGB and FGB, as input. It generates the final segmentation map by integrating the global high-level features from CGB with the fine-grained features extracted by FGB. This synergistic approach of combining multi-modal large models with neural networks has not been extensively investigated. Our experiments show that this method is particularly effective for thyroid nodule segmentation, highlighting the strength of parallel processing between multi-modal models and neural networks before final feature fusion and pixel-wise prediction.

\section{EXPERIMENTS AND DISCUSSIONS}

\begin{table*}[htbp]
	\centering
	\captionsetup{justification=centering, labelformat = simple, labelsep = newline, textfont = sc}
	\caption{COMPARISON WITH OTHER METHODS ON THE COMPREHENSIVE DATASET AND TN3K DATASET}
	\renewcommand{\arraystretch}{1.2}
	\resizebox{\textwidth}{!}{
		\begin{tabular}{cccccccc}
			\toprule
			& & \multicolumn{3}{c}{\textbf{Comprehensive Dataset}} & \multicolumn{3}{c}{\textbf{TN3K Dataset}} \\
			\textbf{Methods} & \textbf{Text} & mIoU(\%) & mDice(\%) & {IoU}$_{\text{FG}}$(\%) & mIoU(\%) & mDice(\%) & {IoU}$_{\text{FG}}$(\%)  \\
			\midrule
			FCN & $\times$ & 82.50 & 88.24 & 67.85 & 76.66 & 83.79 & 60.18 \\
			U-Net & $\times$ & 85.16 & 90.42 & 73.60 & 75.98 & 83.58 & 60.03 \\
			Attention U-Net & $\times$ & 85.97 & 91.04 & 75.06 & 76.65 & 83.98 & 61.09 \\
			ClipSeg & \checkmark & 83.68 & 89.48 & 70.13 & 78.39 & 85.69 & 63.92 \\
			TGANet & \checkmark & 85.22 & 90.94 & 72.70 & 77.89 & 85.11 & 62.13 \\
			LViT & \checkmark & 86.32 & 91.24 & 75.76 & 73.85 & 80.81 & 56.29 \\
			Ours & \checkmark & \textbf{86.85} & \textbf{91.91} & \textbf{76.52} & \textbf{81.29} & \textbf{87.83} & \textbf{69.05} \\
			\bottomrule
		\end{tabular}
	}
	
	\label{tab:tab1}
	\vspace{0.5em}
	\hspace{2em}
	\begin{minipage}{\textwidth}
		\footnotesize  "Text" indicates that the model is a text-augmented segmentation model, while "m" represents the average metric value over the test set. "FG" refers to the foreground.
	\end{minipage}
\end{table*}

\begin{figure*}[!t]
	
	\begin{center}
		\includegraphics[width=1.0 \textwidth]{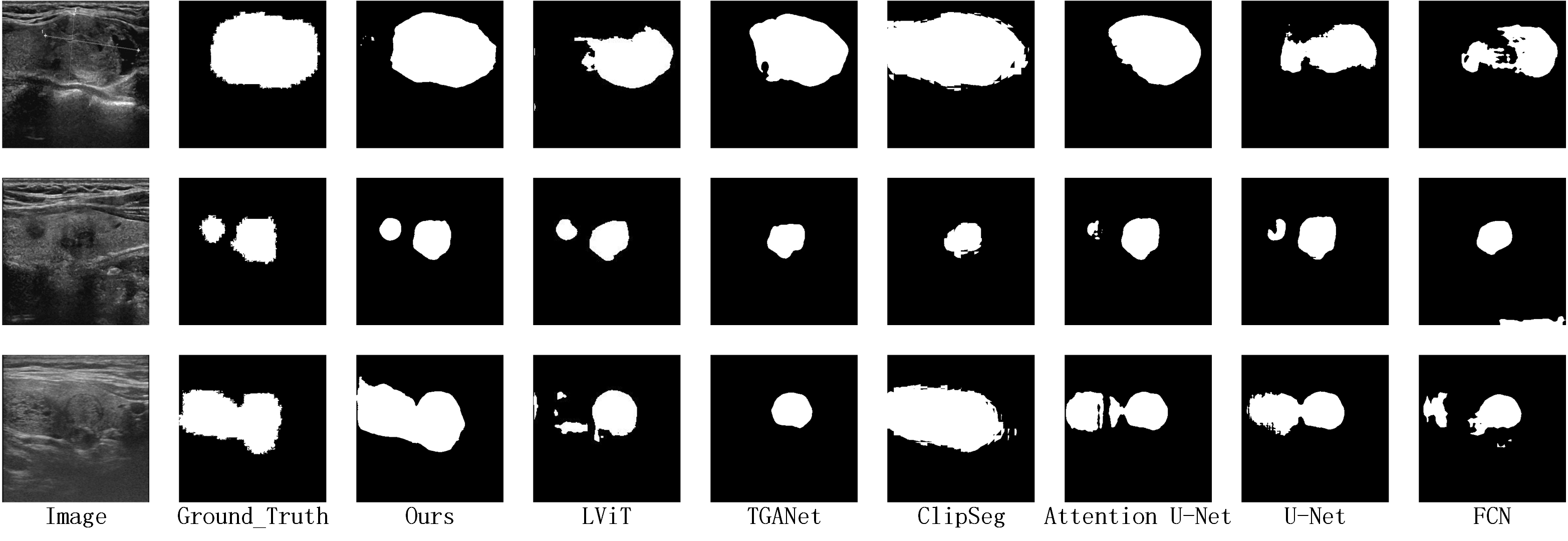}
	\end{center} 
	\caption{Visual comparison of thyroid nodules segmentation results using different methods, from left to right: Input Image, Ground Truth, Ours (CLIP-TNseg), LViT, TGANet, CLIPSeg, Attention U-Net, U-Net, and FCN. } 
	\label{fig:fig2}
	\vspace{-10pt}
\end{figure*}

\subsection{Dataset}
We constructed a large, comprehensive dataset by integrating existing publicly available data and new collections. This dataset includes the DDTI dataset\cite{pedraza2015open}, a newly collected dataset (PKTN) by our team, and additional data sourced from the Internet. Beyond training and evaluation on this dataset, we also validated our model on a separate open-access dataset, TN3K\cite{gong2022thyroid}, to assess its generalization capability.

The PKTN dataset is a newly collected dataset comprising 1,005 ultrasound images sourced from the case database of the Ultrasound Department at Peking University First Hospital, covering cases from 2019 to 2022. The dataset includes benign and malignant thyroid nodules of various sizes, shapes, and echo types. Image dimensions range from (657, 494) to (1680, 1050). All masks were annotated by clinicians with at least 3 years of ultrasound experience, and the annotations were reviewed by a senior clinician with over 5 years of experience. Special care was taken to accurately delineate the nodule boundaries, and in cases where malignant nodules had unclear boundaries, at least two clinicians reached a consensus before finalizing the annotations. The images were acquired in both transverse and sagittal planes following a standardized protocol.

\subsection{Implementation Details}
Our implementation is based on the PyTorch framework \cite{paszke2017automatic}, \cite{paszke2019pytorch}. Images are uniformly cropped to a size of (352, 352) under consideration of object locations, making sure the object remains at least partially visible. In the data augmentation process, we apply a scaling factor of 1.1, random rotations within the range of 0 to 20°, and random translations in the range of [0.99, 1.01]. The training dataset includes images and corresponding textual attributes\cite{wu2020phrasecut}, names, and relationship descriptions. We train the model on an RTX 4080 GPU using the Adam optimizer with a learning rate of 1×$10^{-4}$. The learning rate is adjusted with a cosine annealing strategy, capped at a maximum iteration count, with a minimum learning rate of 1×$10^{-6}$. Training is conducted for 20,000 iterations with a batch size of 16, using Dice loss as the loss function. We employ PyTorch's automatic mixed precision (AMP) strategy to enhance training efficiency, and validation is performed every 500 iterations. The average runtime of CGB, FGB, and PH is 7ms, 8ms, and 1ms, respectively.

\subsection{Performance Comparisons}
We compare our proposed CLIP-TNseg model with established models in medical image segmentation. Among them, ClipSeg\cite{luddecke2022image}, TGANet\cite{tomar2022tganet}, and LViT\cite{li2023lvit} all utilize both text and image inputs, which is the same as our method. All methods are trained on the same dataset, adhering to the parameters specified in the original literature.

Following the common practice of medical image segmentation, we utilize the Dice coefficient and Intersection over Union (IoU) as performance metrics. Notably, we also include the foreground IoU metric, which focuses exclusively on the segmentation performance of foreground pixels, highlighting the model's ability to detect target objects rather than background areas\cite{cai2023corner}.

As shown in Table \ref{tab:tab1}, our method consistently outperforms other existing methods across all datasets and metrics. On the large comprehensive dataset, our model achieves the best performance with IoU of 86.85\%, Dice coefficient of 91.91\%, and foreground IoU of 76.52\%. Furthermore, in the validation on the separate TN3K dataset, our method also demonstrates significantly better results compared to other approaches. Notice that these results were achieved without any additional training on the TN3K dataset. The quantitative results indicate that our method not only provides superior segmentation accuracy but also retains the strong generalization capabilities of large models.

Fig. \ref{fig:fig2} illustrates the predicted segmentation maps for several samples, clearly showing that CLIP-TNseg's predictions align more closely with the target objects and exhibit greater robustness against challenging morphological variations, particularly in cases where existing models struggle to delineate clear boundaries.

\subsection{Ablation Study}
Based on the comparison results, we believe that the performance improvements are likely due to the synergy between the large multi-modal model (CGB) and the neural network architecture (FGB). The primary structure of the target is processed by the CGB, while the FGB ensures reliable boundaries around the main structure.
To verify this, we conducted an ablation study to ensure that both branches contribute to segmentation accuracy. In this experiment, we replaced the output of each branch with zero tensors to evaluate the segmentation accuracy when only one branch is active. As shown in Table \ref{tab:tab2}, the absence of either branch leads to a decline in performance, confirming the indispensability of both components in the proposed method.

\begin{table}[htbp]
	\centering
	\captionsetup{justification=centering, labelformat = simple, labelsep = newline, textfont = sc}
	\caption{RESULTS OF ABLATION STUDIES}
	\resizebox{\linewidth}{!} {
		\begin{tabular}{c c c | c c c}
			\toprule
			\textbf{CGB} & \textbf{FGB} & \textbf{PH} & \textbf{mIoU(\%)} & \textbf{mDice(\%)} & \textbf{IoU}$_\text{FG}$(\%) \\
			\midrule
			\checkmark &  & \checkmark & 79.83 & 86.86 & 63.79 \\
			& \checkmark & \checkmark & 85.82 & 91.07 & 74.89 \\
			\checkmark & \checkmark & \checkmark & 86.85 & 91.91 & 76.52 \\
			\bottomrule
		\end{tabular}
	}
	\label{tab:tab2}
	
\end{table}

\section{Conclusion}

This letter proposes CLIP-TNseg, a novel approach for thyroid nodule segmentation that integrates multimodal large models with neural network architectures. By combining textual and visual supervision, CLIP-TNseg enhances segmentation accuracy, robustness, and generalization. This method shows great potential in clinical applications and can be extended to other medical image segmentation tasks, highlighting the value of multimodal learning in advancing medical diagnostics.

\clearpage
\bibliographystyle{IEEEtran}
\nocite{*}\bibliography{reference}

\end{document}